# Deep Learning for IoT

Tao Lin
*Amazon*
Seattle, US
paper@Ltao.org

*Abstract*— Deep learning and other machine learning approaches are deployed to many systems related to Internet of Things or IoT. However, it faces challenges that adversaries can take loopholes to hack these systems through tampering history data.

This paper first presents overall points of adversarial machine learning. Then, we illustrate traditional methods, such as Petri Net cannot solve this new question efficiently. After that, this paper uses the example from triage(filter) analysis from IoT cyber security operations center. Filter analysis plays a significant role in IoT cyber operations. The overwhelming data flood is obviously above the cyber analyst's analytical reasoning. To help IoT data analysis more efficient, we propose a retrieval method based on deep learning (recurrent neural network). Besides, this paper presents a research on data retrieval solution to avoid hacking by adversaries in the fields of adversary machine leaning. It further directs the new approaches in terms of how to implementing this framework in IoT settings based on adversarial deep learning.

## I. INTRODUCTION

Machine learning, especially Deep Learning, is increasingly popular not only in daily life, but also in many science disciplines, including Internet of Things or IoT[1]. For example, computer security in terms of IoT network intrusions' detection, and malware identification relies on automatic approaches stemming from deep learning, but those are only two examples of deep learning in IoT security. Whereas, deep learning is effective at average normal cases, such as the well-known example of sorting fish automatically by their inherent features. On the other hand, security is targeted on worst tricky cases[2]. It would be easy to bypass a deep learning based IoT analysis filter through malicious tasks in adversarial settings[3]. An example of this is combining malicious samples with benign files, evading several PDF malware classifiers. Therefore, the safe adoption of deep learning approaches in IoT security settings is an unsolved challenge.

Adversarial deep learning is crucial in life-critical IoT systems[4], such as roadside sign recognition used by autonomous vehicles. To be specific, small nonobvious manipulations in roadside signs can lead to distinct opposite results in specific deep learning methods[5]. Accuracy and sensitivity simultaneously are trade-off in several systems.

This paper will focus on two aspects to implement deep learning in adversarial environments using more robust and feasible approaches. First, this paper will suggest do some research on deep learning's transferability. The second research question is about effective defense against adversaries in deep learning in IoT.

Adversaries can launch transferability attacks through constructing an independent deep learning model to simulate some other models just using the input data and output labels[6], without any insights on the original deep learning models' parameters, even the models' type. Transferability is significant not only in adversarial deep learning, but also in many other deep learning applications. On the one hand, although deep learning has achieved impressive successes in many areas, many details are still unclear. Some simple deep learning approaches have similar results with complicated and computationally expensive deep learning algorithms. An open question is whether or not we can simplify the features or hierarchies in deep learning models through transferability. The other question is that although different deep learning models can generate similar outputs from same inputs, we cannot use same evading techniques to attack different deep learning algorithms. In other words, we can protect deep learning models through these transferability properties.

In addition, compared to many state-of-the-art approaches on evading and positioning deep learning models, there is little research to defend the adversaries. Protecting classifiers through ensemble learning, hiding the classification probability scores, or hiding features, even hiding entire classifiers are not appropriate methods, partly because of deep learning's transferability. I propose that defending the adversaries by leveraging reinforcement learning through adversarial training, which is intentionally generating adversarial examples as part of the training procedure. The main challenge is how to craft relevant adversarial examples to simulate the real settings.

## II. FILTER ANALYSIS IN SOC

This section presents traditional data analysis workflow in IoT security operations center, as shown in Figure 1.
Although there is little work on information retrieval related to cyber security operations centers (SOC), there are many works on how to improve and advance the function of SOC. No specific literature for information retrieval on data filter operations. While, information retraval are studied deeply. Generally, there are two categories— graph-based information retrieval approaches and cyber security operations center.

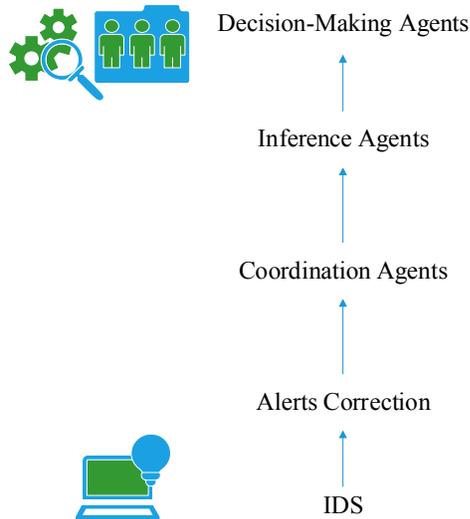

*Figure 1 Traditional data analysis workflow in security operations center*

To some extent, information retrieval and recommendation systems are two sides of one specific coin. By this we mean that a variety of efficient algorithms in recommendation systems can be updated and transformed to the information retrieval. There are some recommendation systems using graph structure. While context-aware information retrieval is a traditional research area, there are increasingly state of the art approaches recently.

Thanks to the increase of software's complexity, Filter analysis confront increasing puzzles, such as:

(1) Undetermined environment. Compared to the past, the loopholes are much more complicated and the GUI events growth dramatically.

(2) Time consuming. Penetration testing mainly depend on random testing and the professional experience of the programmers.

(3) User's unsatisfactory. Though software companies invest a huge amount of human and material resource, it is hard to satisfy the requirements specification.

There are some theory and approaches for Filter analysis in literature. For instance, automatic filter testing can be achieved by a method which they called Observe-Model-Exercise*(* means iteration in the approach). Besides, combinatorial-based criteria can be used in security operations center. There is a filter algorithm for multi-windows analysis. Previously, we used event data graph in the area of GUI testing's coverage rate. However, there is a lack of systematic filter analysis approach based on strict mathematics theory. Therefore, the key to increase the efficiency of filter is that semi-automatic or automatic dynamitic analysis which involves context.

This research proposed a filter method based Petri network to solve the previous three problems and there are three contributions in this research:

(1) Propose and summarize some Petri network properties in filter analysis.

(2) Propose a new automatic analysis approach for six categories IoT deficiencies.

(3) Validate the efficiency of the analysis based Petri network is prior to other traditional methods

We brief how to leverage Petri Net to this area. Firstly, there is a relation explanation between Petri network and IoT analysis, and some significant properties, such as accessibility, boundedness, liveness and strong connectivity. And all the properties are illustrated by a running example, which is a simplified IoT system to explain six categories deficiencies in IoT analysis. The experiments prove that the testing-based Petri network is outperformance than other methods.

Petri network is mainly used in discrete parallel system and the modeling of software. In IoT analysis, Petri network will be migrated to the field. There are some relevant Petri network definitions in IoT analysis. For instance, Petri network is 5-tuple, PN= (P, T, I), In other works, different definitions can be assigned to the other two tuples.

Overall, this section presents Petri Net in IoT analysis. In next section, we will present filter analysis based on deep learning.

### III. DEEP LEARNING BASED RETRIEVAL OF IOT FILTER OPERATIONS

Deep Learning is intriguing far-reaching impacts on information retrieval lately. Compared to a large number of works on image retrieval and natural language processing (NPL), there are tough challenges to other formats (graph, etc.) retrieval through deep learning. By addressing some essential factors in relevance matching, deep relevance matching model (DRMM) can significantly outperform traditional retrieval models. Another model is based on long short-term memory.

Because of the following reasons, deep learning could be used for better IoT data filter retrieval systems.

Firstly, the related methods use of pre-defined similarity measurements [7]. On the other hand, these pre-determined similarity metrics maybe not the most suitable. Deep learning could be leveraged in this scenario to help analysis professional learn how to conquer this.

Secondly, data filter operation retrieval systems are targeted to be used in a variety of uncertainties, such as fault positive. Deep learning could be used in the uncertainties situations.[8] Deep learning, especially recurrent neural networks (RNN) and convolutional neural networks (CNN), is a potential approach, which can be used for data filter operation retrieval in a SOC[8]. We may also conduct research on deep belief networks. In the next sections, we mainly discuss the recurrent neural networks in IoT settings.

### IV. DATA FILTER OPERATION DEEP LEARNING

For data filter operation retrieval[9], many related work uses recurrent neural networks (RNN), since this neural network can be used for data with time sequence.

Instead of rewriting all information, each element in an RNN model updates the current state by adding new information. Accordingly, when an RNN is trained to classify the newly arrived data filter operations, the RNN can be incrementally maintained to incorporate substantial new data triaging knowledge.

But, before training and deploying any RNNs in a SOCs, the SOC should cautiously consider the potential adversaries.

SOC is targeted by adversaries, as shown in Figure 2. That is, the attacker may purposely alter the data to be used for these machine learning settings. The first part of the figure presents all penitential adversary alerts. Traditional machine learning methods can detect these attacks easily. Besides, most approaches can recognize the root events, which is represented by dash lines. However, the most critical attacks are illustrated in the second (bottom) part of the figure. These attacks are drafted by adversaries, which intentionally bypassing machine models. Although these hidden attacks are hard to detect, they still have sequence order. This is the reason we use RNN on this work.

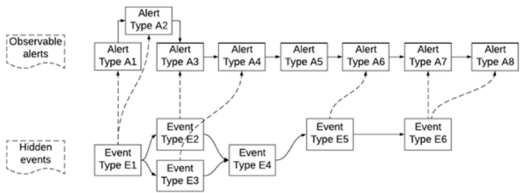

*Figure 2 Data filter operations are being retrieved in adversarial settings*

Recently, a lot of traditional machine learning methods have been hacked by adversaries. The RNNs deployed in a SOC should be resilient to adversarial examples.

The problem is which attacks are more important, accompanied by several trivial network events. In order to do this, most machine learning models use sample events from archive as training data. Then, filter these training data in terms of hierarchies. However, the training data is highly unbalanced. It seems that adversaries can bypassing these models. In this paper, we try to use RNN to overcome these problems.

## V. IoT DATA FILTER MODEL

Cyber security data filter is targeted at determining whether the incoming data sources are worth of further investigation in a timely and quick manner, as shown in Figure 3.

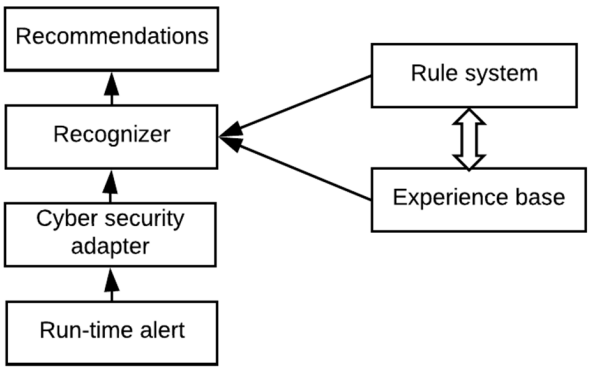

*Figure 3 A retrieval system used by SOC*

To achieve this goal, security analysts usually conduct a sequence of data filter operations to filter malicious network events and to group them according to the potential attack chains bases on the Events Tree, as shown in Figure 4. Therefore, the unit of data filter analysis is a network event. Network events are the data reported by various network monitoring sensors, including SIEM tools and human intelligence agents.

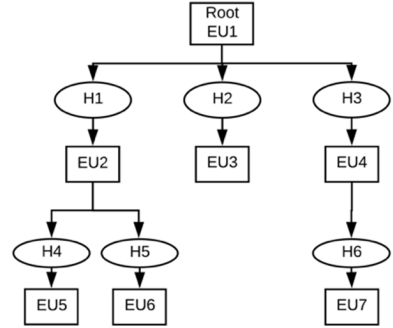

*Figure 4 An Event Tree in IoT SOC*

A *network event* can be abstracted as a multi-tuple of its characteristics,

$$e = <t_{occur}, t_{detect}, type, attack_{prior}, sensor, protocol, ip_{src}, port_{src}, ip_{dst}, port_{dst}, severity, confidence, msg>,$$

where $t_{occur}$ is the date (time) the adversary hacked the IoT system; $t_{detect}$ is the date, when analysist first find the specific adversary task; *type* is the type of these kind of adversary events (e.g., Built, or Deny); $attack_{prior}$ is the analyst using his or her prior experience to determine these tasks; *sensor* is the sensor/agent who first find these tasks; *protocol* is the general networking protocol; $ip_{src}$, $port_{src}$, $ip_{dst}$, $port_{dst}$ are the source IP address, source port number, destination IP address, and destination port number, respectively; *severity* and *confidence* specify the significance of these adversary tasks, respectively; *msg* specifies additional notes for these tuples..

The analyst specified several criteria of the suspicious or correlated network events based on the domain knowledge and experience. Each criterion specifies a constraint on the network event characteristics, so that a data filter operation can select and correlated network events.

## VI. KNOWLEDGE MATCHING AND RULE RELAXATION

Given the rule-based representation, adversaries can mock the similar situation and hack the system based on history data. Therefore, knowledge or experience will not work in these complicated environments.

The problem is: how can we make a limited number of experiences useful for assisting to detect similar events? An experience will not be useful if we do not abstract the particular details. It is significant to retrieve the key parts of an experience and to relax the experience by choosing the portions, which are not too specific.

Conditions with lower priorities can be relaxed. Experience includes specific details, such as the exact time and location of the incident.

The higher the degree to which an experience can be relaxed, the higher the possibility exists that it can be matched against a new situation. Figure 5 shows that the knowledge generated by relaxation form a hierarchy: the most specific knowledge at the bottom while the top is the most relaxed ones.

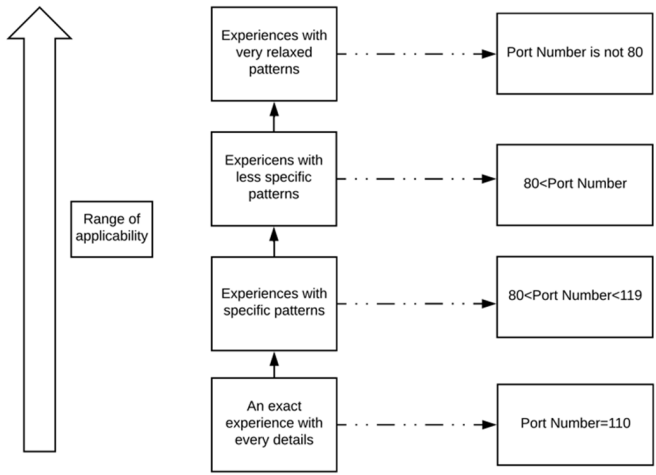

*Figure 5 Experience relaxation levels.*

According to Figure 6, matching based on deep learning is performed on each piece of knowledge through the network. We may use different levels to represent the IoT analysis.

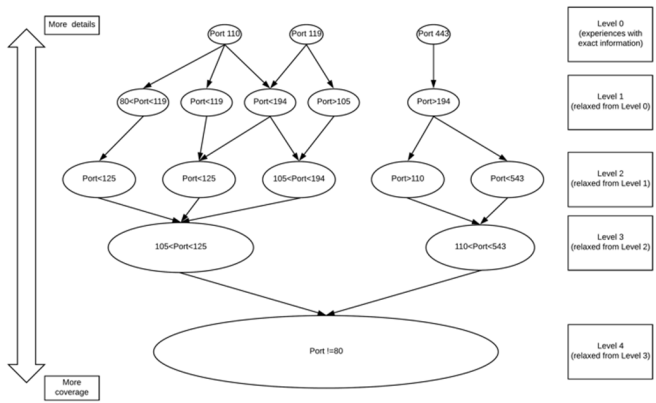

*Figure 6 Hierarchical experience networks.*

## VII. CONCLUSION

This work first presents using Petri Net on SOC retrieval. Then, this work by leveraging RNN to illustrates a new direction for SOC data recommendation. This approach can significantly present the machine learning framework hacked by adversaries.

## VIII. ACKNOWLEDGE

Part of this work is from the author's PhD study[1], before the author joining Amazon. Professor Fu Chen from Central University of Finance and Economics provided many constructive suggestions and perspectives for this work during author's PhD study. Professor Fu Chen and this work were supported in part by National Science Foundation of China under No.61672104.